\documentclass[11pt]{article}
\usepackage{coling2020}
\usepackage{times}
\usepackage{url}
\usepackage{graphics}
\usepackage{graphicx}
\usepackage{amsfonts}
\usepackage{latexsym}
\usepackage{floatrow}
\usepackage{multirow}
\usepackage{multicol}
\newfloatcommand{capbtabbox}{table}[][\FBwidth]

\usepackage[dvipsnames,svgnames,x11names]{xcolor}

\usepackage[markup=underlined]{changes}

\colingfinalcopy 

\title{Improving Commonsense Question Answering by Graph-based Iterative Retrieval over Multiple Knowledge Sources}

\author{Qianglong Chen\textsuperscript{1}, Feng Ji\textsuperscript{2}, Haiqing Chen\textsuperscript{2}, Yin Zhang\textsuperscript{1}\thanks{\textsuperscript{*}Corresponding Author: Yin Zhang} \\
  \textsuperscript{1}College of Computer Science and Technology, Zhejiang University, China \\
  \textsuperscript{2}DAMO Academy, Alibaba Group, China \\
  {\tt \{chenqianglong,zhangyin98\}@zju.edu.cn} \\ {\tt \{zhongxiu.jf, haiqing.chenhq\}@alibaba-inc.com }
  \\
}

\date{}

\begin{document}
\maketitle
\begin{abstract}

In order to facilitate natural language understanding, the key is to engage commonsense or background knowledge. However, how to engage commonsense effectively in question answering systems is still under exploration in both research academia and industry. In this paper, we propose a novel question-answering method by integrating multiple knowledge sources, i.e. ConceptNet, Wikipedia, and the Cambridge Dictionary, to boost the performance. More concretely, we first introduce a novel graph-based iterative knowledge retrieval module, which iteratively retrieves concepts and entities related to the given question and its choices from multiple knowledge sources. Afterward, we use a pre-trained language model to encode the question, retrieved knowledge and choices, and propose an answer choice-aware attention mechanism to fuse all hidden representations of the previous modules. Finally, the linear classifier for specific tasks is used to predict the answer. Experimental results on the CommonsenseQA dataset show that our method significantly outperforms other competitive methods and achieves the new state-of-the-art. In addition, further ablation studies demonstrate the effectiveness of our graph-based iterative knowledge retrieval module and the answer choice-aware attention module in retrieving and synthesizing background knowledge from multiple knowledge sources. 
\end{abstract}

\section{Introduction}
In the past decade, most previous work on question answering systems can be divided into two categories, namely, answering questions within a given context~\cite{rajpurkar2016squad,OpenBookQA2018} or without any  context~\cite{talmor2018commonsenseqa,yang-etal-2018-hotpotqa,sap2019socialiqa}. Although some methods~\cite{Lan2020ALBERT:,zhang2020retrospective} have been reported to exceed human performance on a few metrics, those methods are seldom involved with external knowledge, such as commonsense or background knowledge, which is necessary for better understanding natural language questions, especially in some settings that do not mention the background knowledge. 

To effectively leverage external human-made knowledge graphs, previous methods~\cite{chen2017reading,min2019knowledge,lin2019kagnet,lv2019graph,ye2019align,shen2019multi} adopt a two-stage framework, in which the first stage is to find knowledge facts related to a given question from a wide range of knowledge sources, and then the second stage is to fuse them with the question to predict the answer. \cite{chen2017reading} use a TF-IDF-based Document Retriever to locate relevant paragraphs from Wikipedia documents, and then use a Document Reader to predict the start and end positions of the answer span. \cite{lin-etal-2018-denoising} introduce an additional module of Paragraph Selector to remove noised paragraphs that explicitly do not contain an answer. \cite{das2018multistep} employ a multi-step reasoner to build a new query, then rerank the paragraphs and spot the answer after reading the top paragraph.

Although previous experimental results have proved the effectiveness of incorporating additional knowledge into question answering systems~\cite{speer2017conceptnet,sap2019atomic}, one critical issue still remains to be addressed. As shown in Table \ref{tab:sample}, in ConceptNet, three tail entities, \textbf{midwest}, \textbf{countryside} and \textbf{illinois}, which are also three choices of the example question in CommonsenseQA, are all directly related to the same entity-relation pair $\langle \textbf{farmland}, \textit{AtLocation} \rangle$. If only ConceptNet is given, it is difficult for machine to make the correct choice, since every choice seems correct. We call this issue as multi-value property of knowledge. It is a common phenomenon for question answering system, and the multi-value property of the entity-relation pair will hurt the performance of existing models.

In order to address the above issue, in this paper, we propose a novel question-answering method over multiple knowledge sources. We argue that it is critically important to engage multiple knowledge sources and establish the precise connection between the required background knowledge and the original question as well as choices, to solve the challenge induced by the multi-value property.


We develop our method from three perspectives: 1) We propose a graph-based iterative retrieval module to narrow and refine the potential useful knowledge facts by hidden relations among entities in the question, inspired by \cite{banerjee2019careful,asai2019learning,qi2019answering}. 
2) Different from previous methods only using Wikipedia or ConceptNet, we adopt another extra knowledge source, namely a dictionary, to provide explanations for entities or concepts. Synthesizing entity or concept explanations and iteratively retrieved knowledge facts can help machine precisely distinguish the deceptive answer choices. 3)
Before feeding hidden representations into a linear classifier for final prediction, we introduce an answer choice-aware attention mechanism to compute attention scores between hidden representations of the given question, retrieved knowledge, and candidate choice, which are encoded through a pre-trained language model.


We evaluated our proposed method on the CommonsenseQA dataset, and compared it to other baseline methods. The experimental results show that our method achieves the new state-of-the-art with an improvement of 1.2\% over UnifiedQA~\cite{khashabi2020unifiedqa} in test accuracy. Furthermore, we tested our method using different combinations of external knowledge sources to evaluate the impact of each knowledge source. Furthermore, we carried out ablation studies to evaluate the performance impact of the graph-based iterative knowledge retrieval module and the answer choice-aware attention mechanism on the commonsense question answering task.

The contributions of this paper are summarized as follows:
\begin{itemize}
    \item To improve commonsense question answering, we propose a graph-based iterative knowledge retrieval module, which uses the combination of entities and their potential relations pre-defined in ConceptNet, to find the background knowledge related to the question or its choices from multiple knowledge sources. In addition, we introduce an answer choice-aware attention mechanism to fuse the hidden representation of question, extended knowledge and choices. 
    \item To the best of our knowledge, it is the first time to integrate the explanations of entities or concepts from a dictionary with commonsense or background knowledge to answer questions. Through the extra explanations, our method is able to better distinguish the deceptive answer choices and achieve the new state-of-the-art on the CommonsenseQA dataset.
    \item We also qualitatively and quantitatively evaluated the effectiveness of the proposed graph-based iterative knowledge retrieval module and answer choice-aware attention mechanism in commonsense reasoning setting, and conclude that both of them are useful to improve the performance.
\end{itemize} 

\begin{table}[htb]
    \centering
    \begin{tabular}{l|l|l}
    \hline
    \hline
        \multirow{2}{*}{\textbf{Question}} & \multicolumn{2}{l}{James was looking for a good place to buy \textbf{farmland}.} \\
        & \multicolumn{2}{l}{Where might he look?} \\
    \hline
        \textbf{Choices} & \multicolumn{2}{l}{A: midwest B: countryside C: estate D: farming areas  E: illinois } \\
    \hline
    \hline
        \multicolumn{3}{c}{\textbf{Evidences from different knowledge sources}} \\
        \hline
        \textbf{ConceptNet} & 
        \multicolumn{2}{l}{\textbf{farmland} \small{\textit{AtLocation}} \normalsize \textbf{midwest}} \\ &
        \multicolumn{2}{l}{\textbf{farmland} \small{\textit{AtLocation}} \normalsize \textbf{countryside}} \\ &
        \multicolumn{2}{l}{\textbf{countryside} \small{\textit{IsA}} \normalsize place, \textbf{estate} \small{\textit{RelatedTo}} \normalsize place} \\ &
        \multicolumn{2}{l}{\textbf{countryside} place \small{\textit{RelatedTo}} \normalsize farm} \\ &
        \multicolumn{2}{l}{buy house \small{\textit{HasPrerequisite}} \normalsize  see real \textbf{estate} agent} \\ &
        \multicolumn{2}{l}{\textbf{farmland} \small{\textit{AtLocation}} \normalsize \textbf{illinois}} \\ &
        \multicolumn{2}{l}{\textbf{illinois} \small{\textit{PartOf}} \normalsize \textbf{midwest}} \\
        \hline
        \textbf{Wikipedia} &
        \multicolumn{2}{l}{\textbf{farmland} \textbf{midwest} generally refers to agricultural land, or land currently used for} \\ &
        \multicolumn{2}{l}{the purposes of farming} \\ &
        \multicolumn{2}{l}{five-year pilot project by the \textbf{Countryside} Commission in 1991, the scheme aimed} \\ &
        \multicolumn{2}{l}{to improve the environmental value of \textbf{farmland} throughout England} \\ &
        \multicolumn{2}{l}{\textbf{farmland} in \textbf{Illinois} is valued, as of August 2018, at \$26,000 a hectare} \\
        \hline
        \textbf{Cambridge} &
        \multicolumn{2}{l}{\textbf{farmland}: land that is used for or is suitable for farming} \\ 
        \textbf{Dictionary} &
        \multicolumn{2}{l}{\textbf{midwest}: an area in the US that includes Ohio, Indiana,Michigan, \textbf{Illinois}, Iowa,} \\ &
        \multicolumn{2}{l}{Wisconsin, Minnesota, Nebraska, Missouri, and Kansas} \\ &
        \multicolumn{2}{l}{\textbf{countryside}: land not in towns, cities, or industrial areas,that is either used for } \\ &
        \multicolumn{2}{l}{farming or left in its natural condition} \\ &
        \multicolumn{2}{l}{\textbf{estate}: a large area of land in the country that is owned by a family or an organi-} \\ &
        \multicolumn{2}{l}{zation and is often used for growing crops or raising animals} \\ &
        \multicolumn{2}{l}{\textbf{farming}: the activity of working on a farm or organizing the work there} \\ &
        \multicolumn{2}{l}{\textbf{illinois}: a state in the central US, whose capital city is Spring field and whose } \\ &
        \multicolumn{2}{l}{largest city is Chicago} \\
    \hline
    \hline
    \end{tabular}
    \caption{An example question, its choices, and the relevant knowledge facts retrieved from multiple knowledge sources. (\textit{AtLocation} and \textit{IsA} are potential relations of farmland and its choices.)  }
    \label{tab:sample}
\end{table}

\begin{figure*}[ht!]
    \centering
    \includegraphics[width=1\linewidth]{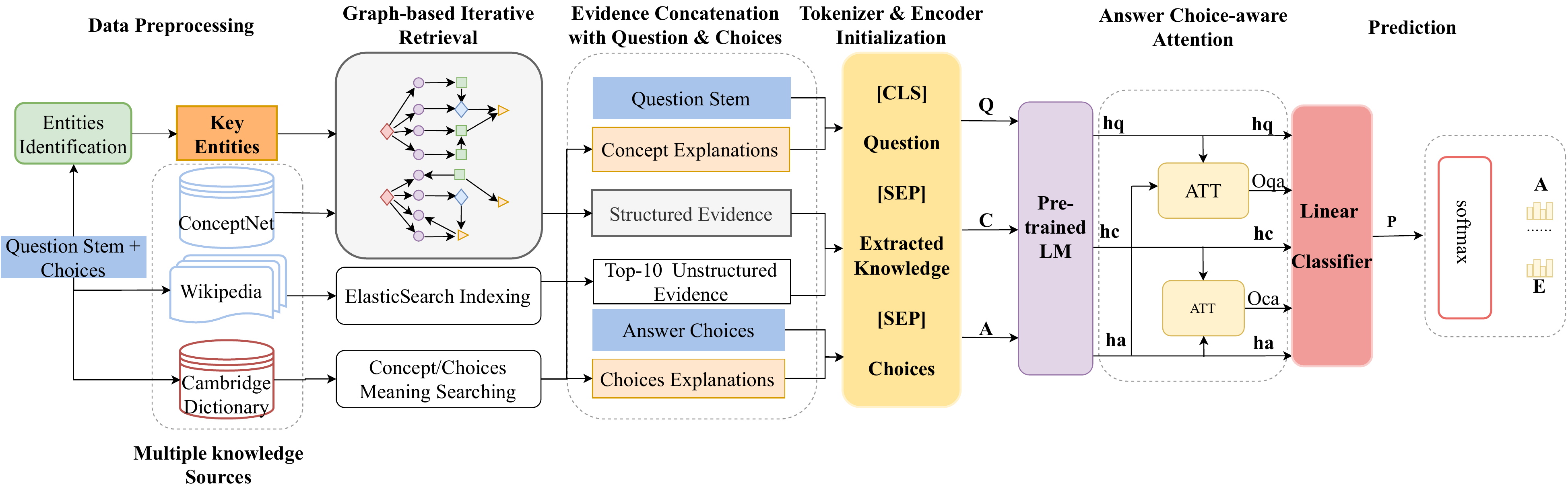}
    \caption{The Overall Architecture. Three knowledge sources, including ConceptNet, Wikipedia and Cambridge Dictionary, are queried to find candidate evidence.
    }
    \label{fig:Model}
\end{figure*}

\section{Approach}

In this section, we present the architecture and details of our approach. As shown in Figure \ref{fig:Model}, our approach consists of four parts: 1) graph-based iterative retrieval module, 2) pre-trained language model based encoding module, 3) answer choice-aware attention mechanism and 4) prediction module. 

In the graph-based iterative knowledge retrieval module, given a question and its choices, we retrieve relevant knowledge facts from multiple knowledge sources. For brevity, we use evidence or knowledge interchangeably to refer to the retrieved knowledge facts. Via the pre-trained language model based encoding module, we encode evidence and its question-answer pair to obtain their hidden representations respectively. In addition, answer choice-aware attention mechanism is designed to compute the attention scores between the question side and its choices side. Entities in both sides are concatenated with additional explanations from the Cambridge Dictionary. Finally, we introduce the linear classifier for specific tasks to predict which answer choice is correct.

\subsection{Problem Formulation}
In this work, we aim to answer questions that require reasoning over multiple evidences from external large-scale knowledge sources. Formally, given a natural language question $Q$ containing $m$ tokens $\{q_1, q_2, ..., q_m\}$, and $n$ choices $\{a_1, a_2, ..., a_n\}$, the objective is to distinguish the right answer from the wrong ones. For evaluation, accuracy is adopted as the metric. An example is shown in Table \ref{tab:sample}. Since the evidence is not provided along with the question and the choices, it naturally requires our method should be able to collect evidences from external knowledge sources and reason over them. Therefore, our method starts from a graph-based iterative retrieval module.

\subsection{Graph-based Iterative Retrieval}
Inspired by the process that people answer questions, instead of the traditional methods, such as n-gram matching or relevant retrieval~\cite{lin2019kagnet,lv2019graph}, we propose a graph-based iterative evidence retrieval method for multiple knowledge sources to retrieve appropriate as evidences.

\begin{figure*}[ht!]
    \centering
    \includegraphics[width=1\linewidth]{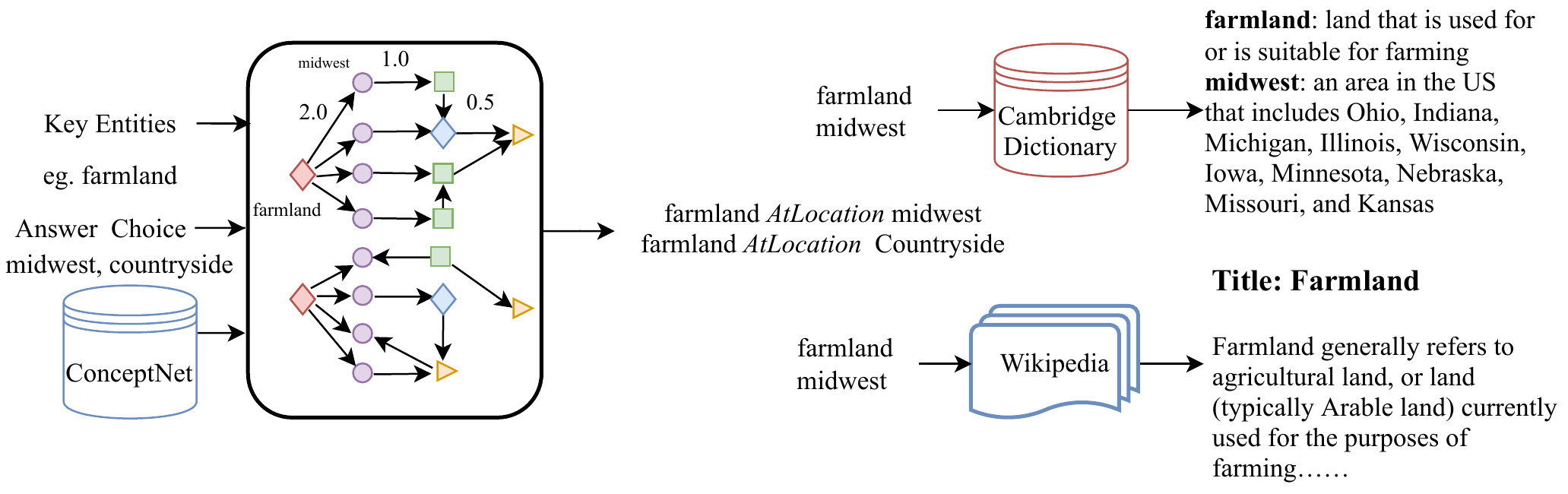}
    \caption{Graph-based Iterative Evidence Retrieval. Colors in each node and line represent different entity types, in which red nodes represent concepts, purple nodes refer to answer choices, green, blue and yellow nodes stand for other type nodes over 2-hops in the graph.}
    \label{fig:retrieval}
\end{figure*}




In the setting of answering multiple-choice questions, the evidence of external knowledge should be related to the concepts of the given question and a choice. As shown in Figure \ref{fig:retrieval}, we first identify the key entities or concepts within a given question and a choice. Specifically, we delete the stop words in questions and use Spacy tool to identify all possible entities. After obtaining the key entities or concepts, we use them to extract the following candidate evidences:
\begin{itemize}
    \item 
    For ConceptNet, we use the question concept and the key entities as initial nodes, including entities in answer choices. According to the type of question, we infer possible relations and use these relations to narrow the scope of knowledge extraction. Specially, for question with token \emph{WHERE}, the possible relations should be related to \emph{AtLocation}. Here, given the type of question, we use rule-based method to infer possible relation. Based on the initial nodes and potential relations, we iteratively retrieve knowledge facts related to the question. In fact, this method is applicable to all structured knowledge graph.
    \item For Wikipedia, similar to~\cite{lv2019graph}, we split Wikipedia documents into sentences, and use ElasticSearch to build indexes. We retrieve the relevant sentences based on the original question and retain the top 10 sentences. For unstructured document text, we can obtain all relevant knowledge in the same way. Since Wikipedia is not a graph by nature,  we construct a graph by treating the elements of the document, including the title, paragraphs, sentences and words, as the nodes, and connecting them to be the edges. Afterward, we use the title and paragraphs to narrow down the scope of the documents. Then we use the paragraph-level candidates and their sentences to build indexes, and retrieve the top 10 relevant sentences.
    \item For the first time, the Cambridge Dictionary is used in the Commonsense QA task. By introducing explanations of entities and concepts, it helps our model distinguish the distracting entities~\cite{talmor2018commonsenseqa}. In order to supplement knowledge, we extract the meaning of the answer choices and the question concept from the dictionary. Entries in the Cambridge Dictionary contain pos tags, pronunciation, explanations, examples, synonyms, etc. In this paper, we only consider the explanations of concepts in a question and answer choices. Although the Cambridge Dictionary is not a graph by nature, we can regard entries in dictionary as sub-graph, which is composed of explanations, examples and synonym. The synonym and examples sentences will connect one entity to another entity. However, we only choose the explanations of related words as extended evidence.
\end{itemize}

Specifically, given the concept of question and the type of question, we infer the required relations and possible keywords in the question context. As shown in the statistics in Table \ref{tab:dataset}, the number of questions with token \emph{WHERE} or \emph{WHAT} accounts for the majority of all samples and they have \emph{AtLocation} and \emph{IsA} relations between the question concept and answer choices. 

For graph-based iterative knowledge retrieval, we use these potential relationships to narrow the scope of external knowledge and regard the concept of question and the choices of answer as initial nodes. As shown in Table \ref{tab:sample}, we present evidences retrieved from ConceptNet, Wikipedia and Cambridge Dictionary, given a question with token \emph{WHERE} and the question concept \emph{farmland}. With the evidences retrieved from multiple knowledge sources, we rank them with BERT fine-tuned on STS-B \cite{cer2017semeval} and select top-10 most relevant evidences for later encoding.


\subsection{Pre-trained Language Model Based Encoding}

After retrieving the knowledge facts related to the question from multiple knowledge sources, we employ the tokenizer to segment the contents of the knowledge, the original questions, and the candidate answers.
Then we feed them into a pre-trained language model for encoding, namely RoBERTa~\cite{liu2019roberta} which is a robust optimized BERT~\cite{devlin2018bert}, since many research work on the pre-trained language models have achieved state-of-art results on a variety of natural language processing tasks. 


Here we denote the encoded question representation as ${Q}=\{q_1,q_2,...,q_m\}$, the encoded choice representation as ${A}=\{a_1,a_2,...,a_n\}$ and the encoded context representation as ${C}=\{c_1,c_2,...,c_k\}$. Specifically, we concatenate the original answer choice  with answer explanation from Cambridge Dictionary as $A$, and concatenate the evidence from Wikipedia and ConceptNet as context $C$. Meanwhile, we concatenate the concept explanation from Cambridge Dictionary with question stem as $Q$.

Formally, the input of pre-trained language model is the concatenation of question $Q$, relevant evidences $C$ and the answer choice $A$.

\begin{equation} 
    h_q = Encoder(Q), h_{a} = Encoder(A), h_c = Encoder(C) \nonumber
\end{equation}

RoBERTa uses a 24-layers transformer architecture, and each block contains a self-attention head and hidden state $H$. We use the last hidden state of RoBERTa as each text encoding representation.

\subsection{Answer Choice-aware Attention Module}
After obtaining the last hidden state from RoBERTa model, for question answering in the downstream task, the previous work usually directly uses linear classifiers to predict the answer.

However, we observed that the linear classifier does not perform well on the retrieved evidences or background knowledge. Therefore, we introduce an answer choice-aware attention mechanism to compute attention scores between the question side $h_q$ and its choices side $h_a$, as well as attention scores between retrieved evidences $h_c$ and answer choices $h_a$ via the standard attention calculation.

\begin{equation}
    O_{qa} = ATT(h_q,h_a), O_{ca} = ATT(h_c,h_a) \nonumber
\end{equation}

\subsection{Prediction}

We concatenate attention reweighted hidden states and pass them through a linear classifier with ReLU \cite{nair2010rectified} to compute the final bidirectional attention vectors for prediction. The formula is as follows:
\begin{equation}
    P(q,a) = Linear(O_{qa} h_q,O_{ca} h_c,h_a)
\end{equation}

\begin{table*}[ht!]
    \centering
    \begin{tabular}{c|cccccc|c}
         \hline
         \hline
         \multirow{2}*{\textbf{Dataset}} & \multicolumn{6}{c|}{\textbf{Question Type}} & \multirow{2}*{\textbf{Total}} \\
         \cline{2-7}
         & how & what & where & when & why & others & \\
         \hline
         train  & 227 & 6142 & 2885 & 67 & 243 & 177 & 9741 \\
         dev &  30 & 784 & 345 & 10 & 35 & 17 & 1221  \\
         test &  31 & 728 & 321 & 9 & 25 & 26 & 1140 \\
         \hline
         \hline
    \end{tabular}
    \caption{Overall statistics of question types on the train, dev and test sets of CommonsenseQA. Question types are identified with heuristic rules, thus the counts on question types are not accurate since a few questions contain more than one trigger word.}
    \label{tab:dataset}
\end{table*}

\section{Experiments}
\subsection{Dataset}
\paragraph{CommonsenseQA}~\cite{talmor2018commonsenseqa} is a new multiple-choice question answering dataset that requires different types of commonsense knowledge to predict the correct answers. It is challenging for existing pre-trained language models such as ALBERT, RoBERTa, and T5. Here is an example in Table \ref{tab:sample}. We use official split CommonsenseQA  dataset. As shown in Table \ref{tab:dataset}, it contains 12,102 questions with one correct answer and four candidate answers, including 9,741 for training, 1,221 for development and 1,140 for test. Moreover, the answer choices often have similar relations in ConceptNet. Therefore, multiple knowledge sources must be utilized to distinguish the choices.

\subsection{Knowledge Sources}
We choose ConceptNet\footnote{https://conceptnet.io/} and Wikipedia\footnote{https://dumps.wikimedia.org/enwiki/} as the structured knowledge graphs and unstructured documents, and choose Cambridge Dictionary\footnote{https://dictionary.cambridge.org/us/} to provide vocabulary explanation.

\paragraph{ConceptNet}
ConceptNet~\cite{speer2017conceptnet} is a freely-available semantic network, designed to help computers understand the meanings of words that people use. It is one of the most largest structured knowledge bases knowledge from other crowd sourced resources, expert-created resources, and games with a purpose. Because of its huge size, we choose it as the representative structured knowledge source. 

\paragraph{Wikipedia}
Wikipedia is a free online encyclopedia, created and edited by volunteers around the world and hosted by the Wikimedia Foundation. We choose Wikipedia 22-May-2020 version as the unstructured knowledge source and use wiki extractor to extract documents and split them into sentences.

\paragraph{Cambridge Dictionary}
The Cambridge Advanced Learner's Dictionary was first published in 1995. The dictionary has over 140,000 words, phrases, and meanings. We choose Cambridge Dictionary as the dictionary knowledge source to obtain words or concept meanings for model understanding.

\subsection{Baselines}
\paragraph{Language Models} Language models include BERT-large~\cite{devlin2018bert}, XLNET-large~\cite{yang2019xlnet}, RoBERTa~\cite{liu2019roberta}, ALBERT~\cite{Lan2020ALBERT:} and T5~\cite{raffel2019exploring}. These models adopt pre-trained language models to fine-tune on the training data and make predictions on test dataset without knowledge extraction.
\paragraph{KagNet} KagNet~\cite{lin2019kagnet} utilizes external, structured commonsense knowledge graphs to perform explainable inferences. It grounds a question-answer pair from the semantic space to the knowledge-based symbolic space as a schema graph, a related sub-graph of external knowledge graphs.
\paragraph{AristoBERTv7}
AristoBERTv7\footnote{https://leaderboard.allenai.org/arc/submission/bk5snmbvhqhm94h7heag} starts from the RoBERTa-large model, it is fine-tuned on RACE training set and further fine-tuned on a combination of other science QA datasets~\cite{OpenBookQA2018,clark2018think}, finally fine-tuned on CommonsenseQA training set using the model with best dev score. For each question + answer choice, AristoBERTv7 retrieves the 10 best sentences (by TF-IDF) which have non-stopping words overlapped with both question and answer choice. Then they concatenate the sentences in an increasing order of matched score and feed them into RoBERTa as the context part of the input besides question tokens and answer tokens. Moreover, it truncates the context tokens from the left if the number of words is greater than the "max tokens" setting of 256. 
\paragraph{DREAM}
DREAM \footnote{https://paper.dropbox.com/doc/Model-Details-LWgdNCGtzW7Q1brYFKihD} adopts XLNet-large as the baseline and extracts evidence from Wikipedia. They first use ElasticSearch to build indexes for Wiki docs and find top-10 sentences using BM25. The query is a question + answer choice. Then, they concatenate them to fine-tune XLNet-large cased model. 
\paragraph{RoBERTa + KG}
RoBERTa + Knowledge Extraction(KE) \footnote{https://github.com/jose77/csqa/blob/master/desc.md}, RoBERTa + Information Retrieval(IR) \footnote{https://1drv.ms/b/s!Aq1PIOBthMoKblvGqds3CzR451k?e=Yg6P94} and RoBERTa + CSPT \footnote{https://gist.github.com/commonsensepretraining/} adopt RoBERTa as the baseline and utilize the evidences from Wikipedia, search engine and Open Mind Common Sense (OMCS), respectively.
RoBERTa + IR first fine-tunes RoBERTa large model on the RACE dataset by concatenating the hidden representation of passages, question and answer choices. Then they retrieve context information for each question of CommonsenseQA through the search engine, and further fine-tune on train data. 
RoBERTa + KE first retrieves 10 best sentences from wiki docs for each combination of question and the choice as the context. Then they fine-tune the pre-trained RoBERTa model on CommonsenseQA dataset. RoBERTa + CSPT first trains a generation model to generate synthetic data from ConceptNet. Then they build the commonsense pre-trained model by fine-tuning RoBERTa-large model on the synthetic data and Open Mind Common Sense (OMCS) corpus. The final model is fine-tuned from the pre-trained model on CommonsenseQA.
\paragraph{FreeLB-RoBERTa}
FreeLB \cite{Zhu2020FreeLB:} promotes higher invariance in the embedding space, by adding adversarial perturbations to word embeddings and minimizing the resultant adversarial risk inside different regions around input samples, which is applied to transformer-based models for natural language understanding and commonsense reasoning tasks.

\paragraph{XLNET + Graph Reasoning}
 XLNET + Graph Reasoning \cite{lv2019graph} extract evidences from both structured knowledge base and Wikipedia plain texts and construct graphs for both sources to obtain the relational structures of evidence. Based on these graphs, they propose a graph-based approach consisting of a graph-based contextual word representation learning module and a graph-based inference module for reasoning. However, they did not consider extending the entities with words explanations and meanings from dictionary.
 
\paragraph{ALBERT + Path Generator} 
\cite{wang2020connecting} propose a multi-hop knowledge path generator to generate structured evidence dynamically according to the question. They use a pre-trained language model as the backbone, leveraging a large amount of unstructured knowledge stored in the language model to supplement the incompleteness of the knowledge base.

\paragraph{UnifiedQA}
UnifiedQA \cite{khashabi2020unifiedqa} presents a new unified pre-trained language model which is pre-trained using three types of language modeling tasks: unidirectional, bidirectional, and sequence-to-sequence prediction, and employs a shared transformer network and specific self-attention masks to control what context the prediction conditions on.

\begin{table}[ht!]
    \centering
    \begin{tabular}{l|c|c}
    \hline
    \hline
    \textbf{Model name} & \textbf{Dev Accuracy(\%)} & \textbf{Test Accuracy(\%)} \\
    \hline
    \hline
    KagNet & 64.46 & 58.9 \\
    BERT + OMCS & 68.8 & 62.5 \\
    AristoBERTv7 & - & 64.6 \\
    DREAM & 73.0 & 66.9 \\
    RoBERTa + KE & 77.5 & 68.4 \\
    RoBERTa + CSPT & 76.2 & 69.6 \\
    RoBERTa & 78.4 & 72.1 \\
    RoBERTa-IR & 78.9 & 72.1 \\
    FreeLB-RoBERTa & 78.81 & 72.2 \\
    XLNET + Graph Reasoning & 79.3 & 75.3 \\
    ALBERT + Path Generator & 78.42 & 75.6 \\
    PEAR & 78.42 & 76.1 \\
    MHGRN & - & 76.5\\
    ALBERT (ensemble model) & 83.7 & 76.5 \\
    T5 & - & 78.1 \\
    ALBERT + Path Generator (ensemble model) & - & 78.2\\
    UnifiedQA & - & 79.1 \\
    
    \hline
    \hline
    \textbf{Human Performance} & - & 88.9 \\
    \hline
    \hline
    \textbf{Our Model} & \textbf{87.4} & \textbf{80.3} \\
    \hline
    \hline
    \end{tabular}
    \caption{Results on CommonsenseQA development and blind test dataset. Dev Accuracy denotes accuracy on development set. Test Accuracy denotes accuracy on test set.}
    \label{tab:result_1}
\end{table}

\subsection{Experimental Setting}
We select RoBERTa-large as the pre-trained model which uses a 24-layers transformer architecture. We set the max update step to 6000, warmup update step to 150, and max length to 512. We set droput to 0.1. Meanwhile, we use the Adam\cite{kingma2014adam} as the optimizer and adopt cross-entropy loss as our loss function. In our best model on the development dataset, we set the batch size to 4 and learning rate to 1e-5.

\subsection{Experimental Results and Analysis}
The results on CommonsenseQA development dataset and blind test dataset are shown in Table 3. Our model achieves the best performance on both datasets.

Our model performs better than language models such as BERT-large \cite{devlin2018bert}, XLNET-large \cite{yang2019xlnet}, RoBERTa \cite{liu2019roberta}, ALBERT\cite{Lan2020ALBERT:} and T5 \cite{raffel2019exploring}, which indicates that the external knowledge facts retrieved from multiple sources can provide more information to help the system answer question correctly and break limitations of the existing pre-trained language models.
Meanwhile, our model performs better than XLNET + Graph Reasoning, which uses ConceptNet \cite{speer2017conceptnet} and Wikipedia as their knowledge sources. It brings an improvement of $5\%$, which indicates that leveraging the explanations of entities or concepts benefits a lot to this task when compared to methods treating ConceptNet or Wikipedia as the only external knowledge source.

Thus, conclusions have come up that our graph-based iterative retrieval module can effectively work over multiple knowledge sources and entity explanation indeed contributes to significantly boosting the performance of multiple choice question answering. Although it remains a big gap to human performance, our method moves ahead and achieves the new state-of-the-art.

\subsection{Ablation Study}

    
    

    
    

In this section, we perform ablation studies on the test dataset to evaluate the impacts of different components and knowledge sources employed in our method.

We select RoBERTa without evidence as the baseline. In the baseline, we simply concatenate the question, answer choices into RoBERTa and adopt the commonsense reasoning module for prediction.

\begin{table*}
\begin{floatrow}
\capbtabbox{
 \begin{tabular}{l|c}
        \hline
         \hline
         \textbf{Model} & \textbf{Test Accuracy} \\ 
         \hline
         \hline
         RoBERTa (w/o KG) & 72.1 \\
         + ConceptNet & 74.4 \\
         + Wiki & 73.0 \\
         + ConceptNet + Wiki & 75.3 \\
         + Cambridge Dictionary & 77.6 \\
         + All & 80.3 \\ 
         \hline
         \hline
    \end{tabular}
}{
 \caption{Ablation studies of knowledge sources on CommonsenseQA.}
\label{tab:ablation1}
}
\capbtabbox{
 \begin{tabular}{l|c}
        \hline
         \hline
         \textbf{Model} & \textbf{Test Accuracy} \\ 
         \hline
         \hline
         RoBERTa (w/o Modules) & 72.1 \\
         + Graph-based Iterative & \multirow{2}{*}{77.4} \\
         Retrieval &  \\
         + Answer choice-aware Attention  & 75.8 \\
         + Both & 80.3 \\
         \hline
         \hline
    \end{tabular}
}{
 \caption{Ablation studies of Model architecture on CommonsenseQA.}
    \label{tab:ablation2}
}
\end{floatrow}
\end{table*}

By extracting evidence from multiple knowledge sources, we can obtain an overall improvement of 8.2\% over the baseline. All experimental results are shown in Table \ref{tab:ablation1}. Compared to the baseline, the evidence from ConceptNet brings a gain of 2.3\% while 0.9\% from the unstructured Wikipedia document. When leveraging the evidence from both ConceptNet and Wikipedia, a moderate gain of 3.2\% can be obtained. However, if only using the explanations from Cambridge Dictionary, the improvement is bigger than 5.5\%. This proves that the explanations from Cambridge Dictionary are more helpful to clarify the knowledge with multi-value property, thus finally improve the accuracy. After using all evidence from three heterogeneous knowledge sources, the final accuracy is 80.3\%.  

In the second ablation study, we extract evidence from multiple knowledge sources iteratively. In commonsense reasoning part, we incorporate the knowledge evidence with question answer tokens and use answer choice-aware attention to compute the attention coefficients. Then, we  integrate the attention score with hidden states and use linear classifier for prediction.

As shown in Table \ref{tab:ablation2}, the graph-based iterative retrieval module can bring a gain of $5.3\%$, while $3.7\%$ by adding the answer choice-aware attention mechanism. It means that both of them can find out the effective evidence along with explanations from multiple knowledge sources.

\subsection{Case Study}
In this section, we demonstrate a case to show that our model can utilize the knowledge sources to answer questions. As shown in Table \ref{tab:sample}, the question is “James was looking for a good place to buy farmland. Where might he look?” and the answer is “midwest”. The facts are retrieved from ConceptNet, Wikipedia, Cambridge Dictionary shown in Figure \ref{fig:retrieval}. 

The evidence from ConceptNet shows farmland has somewhat relationships with midwest, countryside and illinois, which are also the answer choices in the candidate answers. The evidence from Wikipedia shows farmland often appears with illinois. Just based on these evidence, we may choose illinois. However, the explanation of midwest in Cambridge Dictionary is an area in US, including Ohio, Indiana, Michigan, Illinois, etc. The explanation of farmland is the land used for or suitable for farming. Going back to the example,
James was looking for a good place to buy farmland. It means that he might look for such a place in the midwest rather than the countryside. Therefore, midwest is a more suitable answer than other choices.


\section{Related Work}

\subsection{Benchmarks}
Datasets like SQuAD \cite{rajpurkar2016squad}, WikiQA \cite{yang2015wikiqa}, TriviaQA \cite{joshi2017triviaqa}, CoQA \cite{reddy2019coqa}  have gained enormous attention over the past few years. Since the answers are presented within context, the questions from these datasets are easy to solve.
The language models \cite{liu2019roberta} trained on huge amount of data have been able to compete with humans on datasets like OpenBookQA \cite{OpenBookQA2018}.
Recent benchmarks such as CommonsenseQA  \cite{talmor2018commonsenseqa} focuse on factual and physical commonsense derived from ConceptNet.
\cite{rajani-etal-2019-explain} explore adding human-written explanations to solve the problem. \cite{lin2019kagnet} propose to extract evidence from ConceptNet to study this problem. 

In this paper, we deal with datasets which not only require external facts but also need commonsense knowledge to predict the correct choices like CommonsenseQA.

\subsection{Information Retrieval for Knowledge Facts}
There are many research work on answering questions with external domain-specific knowledge bases or generating responses containing domain-specific attributes and entities for task-oriented dialogues~\cite{hao-etal-2017-end,das-etal-2017-question,madotto-etal-2018-mem2seq,lin-etal-2019-task}. However, in this paper, we aim at answering questions over open-domain knowledge, such as CommonsenseQA.
In such problems, most of the past proposed  methods\cite{lin2019kagnet,lv2019graph,min2019knowledge,ye2019align,shen2019multi} use traditional information retrieval methods such as n-gram matching and TF-IDF.
\cite{lin2019kagnet} propose a straightforward approach with two key components, namely soft matching through lemmatization and finding a path for sub-graph matching. \cite{lv2019graph} adopt ElasticSearch, a search engine toolkit, to index the Wikipedia sentences and rank by the matching scores between the query and all Wikipedia sentences. \cite{banerjee2019careful} firstly propose to extract knowledge by retrieving top 10 relevant facts from the knowledge source through a pre-trained model on STS-B dataset \cite{cer2017semeval}. 

As retrieved knowledge is not precise enough to satisfy the requirements of open questions, traditional search technologies or semantic matching algorithms do not work well on such datasets. \cite{feldman-el-yaniv-2019-multi} introduce a multi-hop paragraph retrieval method for open-domain question answering. Inspired by their work, we consider the knowledge retrieve as a open-domain information retrieval and propose a graph-based iterative retrieval to find evidences from multiple knowledge sources.

\section{Conclusion}
In this paper, we deal with the multi-choice question answering task which requires background knowledge or commonsense. We propose a novel question-answering method by exploring how to efficiently integrate multiple knowledge sources, i.e. ConceptNet, Wikipedia and the Cambridge Dictionary. Firstly, we propose a novel graph-based iterative knowledge retrieval module to iteratively retrieve concepts and entities related to a given question and its choices. In addition, we propose an answer choice-aware attention mechanism to fuse all hidden representations encoded by a pre-trained language model. We conducted experiments on the CommonsenseQA dataset and the experimental results show that our method significantly outperforms other competitive methods in accuracy. Further ablation studies show the effectiveness of graph-based iterative knowledge retrieval module and answer choice-aware attention module in retrieving and synthesizing background knowledge from multiple knowledge sources. In the future, we will extend our method to deal with the open-domain question answering tasks that require the external background knowledge.
\section*{Acknowledgements}
This work was supported by the NSFC projects (No. 61402403, No. 62072399, No. U19B2042), Chinese Knowledge Center for Engineering Sciences and Technology, MoE Engineering Research Center of Digital Library, Alibaba-Zhejiang University Joint Institute of Frontier Technologies, and the Fundamental Research Funds for the Central Universities. 

\bibliographystyle{coling}
\bibliography{coling2020}

\end{document}